\documentclass{article}

\usepackage[numbers]{natbib}

 \usepackage[preprint]{neurips_2025}

\usepackage[utf8]{inputenc} 
\usepackage[T1]{fontenc}    
\usepackage{hyperref}       
\usepackage{url}            
\usepackage{booktabs}       
\usepackage{amsfonts}       
\usepackage{nicefrac}       
\usepackage{microtype}      
\usepackage{xcolor}         

\usepackage{graphicx}
\usepackage{subfig}
\usepackage{amsmath}
\usepackage{wrapfig}
\usepackage{pifont}
\usepackage{dsfont}

\title{Maximizing Confidence Alone Improves Reasoning}

\author{%
  Mihir Prabhudesai\thanks{Equal contribution.} \\
  Carnegie Mellon University\\
  \And
  Lili Chen\textsuperscript{*} \\
  Carnegie Mellon University\\
  \And
  Alex Ippoliti\textsuperscript{*} \\
  Carnegie Mellon University\\
  \AND
  Katerina Fragkiadaki \\
  Carnegie Mellon University\\
  \And
  Hao Liu \\
  Carnegie Mellon University \\
  \And
  Deepak Pathak \\
  Carnegie Mellon University\\
}

\begin{document}

\maketitle

\begin{abstract}
  Reinforcement learning (RL) has enabled machine learning models to achieve significant advances in many fields. Most recently, RL has empowered frontier language models to solve challenging math, science, and coding problems. However, central to any RL algorithm is the reward function, and reward engineering is a notoriously difficult problem in any domain. In this paper, we propose \textbf{RENT:} \textbf{R}einforcement Learning via \textbf{Ent}ropy Minimization -- a fully unsupervised RL method that requires no external reward or ground-truth answers, and instead uses the model's entropy of its underlying distribution as an intrinsic reward. We find that by reinforcing the chains of thought that yield high model confidence on its generated answers, the model improves its reasoning ability. In our experiments, we showcase these improvements on an extensive suite of commonly-used reasoning benchmarks, including GSM8K, MATH500, AMC, AIME, and GPQA, and models of varying sizes from the Qwen, Mistral, and Llama families. The generality of our unsupervised learning method lends itself to applicability in a wide range of domains where external supervision is unavailable. Website: \url{https://rent-rl.github.io/}.
\end{abstract}

\section{Introduction}

Imagine you're taking an exam. Once it begins, no new information is available and no external help can be sought. With only your own reasoning to rely on, how do you tackle a difficult problem? You might make an initial attempt, assess your confidence in the answer, and revise your reasoning until you feel sufficiently certain. Of course, confidence is not a guarantee of correctness -- but in the absence of feedback, it is often the only intrinsic signal we have to guide further thought. In such settings, humans tend to optimize for confidence, or equivalently, to reduce uncertainty.
\begin{figure}[h!]
    \centering
    \includegraphics[width=0.98\textwidth]{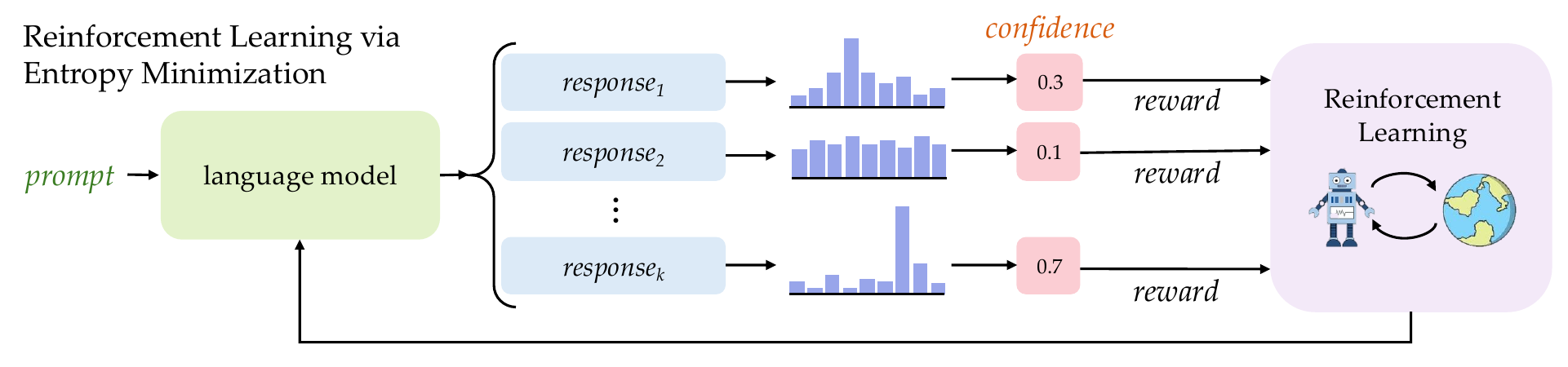}
    \caption{Overview of RENT: Reinforcement Learning via Entropy Minimization. For each response, we use the model's underlying confidence (negative entropy) as a reward for reinforcement learning. This enables the model to learn without any external reward or ground-truth answers.}
\end{figure}
In machine learning, uncertainty is commonly quantified via entropy -- a measure of how peaked or diffuse a probability distribution is. Language models output distributions over tokens, and the entropy of these distributions reflects the model's confidence: lower entropy implies more confident predictions. Yet despite the growing use of language models in reasoning tasks, current approaches to improvement still rely heavily on external supervision, rewarding models based on correctness with respect to ground-truth labels ~\cite{guo2025deepseek,shao2024deepseekmath}. This dependence is often impractical, particularly in real-world or open-ended scenarios where supervision is scarce or unavailable.

To address this, we propose \textbf{RENT:} \textbf{R}einforcement Learning via \textbf{Ent}ropy Minimization -- a fully unsupervised reinforcement learning method that improves reasoning performance by using the model’s own confidence as a reward. Specifically, we define the reward as the negative entropy of the model’s predicted token distributions. This signal is dense, general, and easy to compute, requiring no ground-truth answers. Importantly, not all parts of the response contribute equally to final performance. Through empirical analysis, we find that minimizing entropy over tokens near the end of the reasoning chain, especially those corresponding to the final answer, correlates most strongly with improved accuracy. In contrast, early tokens in the response show little correlation. This suggests that as the model approaches its final answer, it increasingly relies on its own confidence to guide reasoning, so encouraging confidence in these final steps is key to improving overall performance.

We demonstrate RENT’s effectiveness across diverse reasoning benchmarks, including GSM8K~\citep{cobbe2021training}, MATH500~\citep{hendrycks2021measuring,lightman2023let}, AMC and AIME~\citep{li2024numinamath}, and GPQA~\citep{rein2024gpqa}. Our method scales across model families (Qwen, Mistral, and Llama) and sizes and consistently improves performance.

\section{Related Work}
\subsection{Reinforcement Learning for Reasoning}
Initially, reinforcement learning (RL) for language models was mostly used for learning from human preferences~\citep{christiano2017deep} and, traditionally, the RL optimization was done with algorithms such as PPO~\citep{schulman2017proximal}. With the capabilities of language models continuing to improve, researchers have begun to explore the possibility of using RL to improve the performance of language models on reasoning tasks such as math~\citep{cobbe2021training,hendrycks2021measuring,li2024numinamath}, science~\citep{he2024olympiadbench,rein2024gpqa}, or coding~\citep{li2022competition, codealpaca} problems. In these settings, the model is prompted to generate a chain-of-thought~\citep{wei2022chain} and final answer, and receives a reward based on how closely its final answer matches the ground-truth answer. These efforts present RL as an alternative to search-based approaches to chain-of-thought reasoning such as Tree of Thoughts~\citep{yao2023tree} and Graph of Thoughts~\citep{besta2024graph}. Related lines of work include training a reward model to give feedback for every step in the chain of thought, and training RL models to encourage self-correcting behaviors in language models. Examples of RL methods in this space include \citet{zelikman2022star,singh2023beyond,kumar2024training,qu2024recursive,uesato2022solving,lightman2023let,wang2023math}. At scale, DeepSeek~\citep{guo2025deepseek, shao2024deepseekmath} proposed an open-source model that showed OpenAI o1~\citep{jaech2024openai}-level reasoning by performing RL in this manner, using a new algorithm GRPO~\citep{shao2024deepseekmath}.

\subsection{Confidence and Calibration}
Confidence measures quantify how certain a model is that its generated output is correct~\citep{yoon2025reasoning,spiess2024calibration}. In order to evaluate the confidence of machine learning models, it is necessary also to discuss \textit{calibration}~\citep{kalai2024calibrated,virk2024enhancing} - i.e., how aligned those confidences are with actual correctness. As language models are increasingly trusted to make important decisions, providing users with a reliable confidence measure would be useful in many situations~\citep{geng2023survey,manakul2023selfcheckgpt,varshney2023stitch,hou2023decomposing,jiang2023active,han2024towards,spiess2024calibration}. As such, researchers have developed various confidence metrics for modern deep learning models and studied the extent to which they are calibrated. These include both methods that assume access to the model's weights~\citep{gupta2024language,kadavath2022language,xu2024sayself,spiess2024calibration} and methods that estimate confidence via prompting alone~\citep{xie2024survey,geng2023survey,tian2023just,xiong2023can,yang2024verbalized}. In our paper, we use the model's confidence to iteratively improve its own performance via reinforcement learning. 

\subsection{Test-Time Adaptation}
Test-time adaptation is where a model is updated using data from the test distribution, without access to ground-truth labels. The goal is to improve performance in scenarios where there is a distribution shift between training and testing environments. Methods for adapting without labels include normalization techniques that recalibrate feature statistics at test time ~\citep{quinonero2008covariate,sun2017correlation,maria2017autodial,schneider2020improving,nado2020evaluating}. The most relevant work to ours is Tent~\citep{wang2020tent}, which performs entropy minimization on model predictions during test time. This approach assumes that predictions on test data should be low in entropy if the model is well-adapted to the new distribution. Tent builds on earlier work that uses entropy minimization as a regularization strategy in semi-supervised learning~\citep{grandvalet2004semi,lee2013pseudo,berthelot2019mixmatch} and domain adaptation contexts~\citep{maria2017autodial,shu2018dirt,saito2019semi}, where encouraging confident predictions has proven effective for improving generalization. Recently, TTRL~\citep{zuo2025ttrl} proposed test-time reinforcement learning using majority voting as a reward. Compared to entropy, majority voting is a sparse reward and much less general; for example, it cannot be applied to long-form free-response questions.

\subsection{Unsupervised Reinforcement Learning}

Unsupervised RL trains agents using intrinsic rewards like novelty, entropy, or mutual information, enabling skill acquisition without extrinsic feedback. Prior methods include ICM and RND for prediction error \cite{pathak2017curiosity,burda2018exploration}, APT \cite{liu2021behavior} and ProtoRL \cite{yarats2021reinforcement} for entropy maximization, and DIAYN, APS, and SMM for skill discovery via mutual information \cite{eysenbach2018diversity,liu2021behavior,kim2023variational}. An interesting observation is that while exploration methods primarily maximize entropy, we instead minimize it by reinforcing high-confidence outputs, and find that for language models, this leads to better reasoning performance without any external supervision. 

\section{Method}
\subsection{Reinforcement Learning for Language Models}
The goal of reinforcement learning (RL) is to train a policy which generates actions that maximize the cumulative expected reward. In the context of modern language models, the policy $\pi$ is a language model and the actions $y_{\text{pred}}$ are sampled from the distribution over a discrete vocabulary. The task is formulated as a one-step RL problem in which the model generates $y_{\text{pred}} = \pi(x)$, where $x$ is sampled from the dataset $\mathcal{D} = \{(x, y_{\text{target}})\}$, and receives some reward for the generation. Typically, the ground-truth answer $y_{\text{target}}$ is used to give the model a reward $r = \mathcal{R}(y_{\text{target}},y_{\text{pred}})$. One reward function which is currently used is simple string matching, where $\mathcal{R}(y_{\text{target}},y_{\text{pred}}) = \mathds{1}\{y_{\text{target}}=y_{\text{pred}}\}$. Our work focuses on instead doing \textit{unsupervised} reinforcement learning, which does not require external supervision for the reward. Specifically, $y_{\text{target}}$ is not used in the reward $r=\mathcal{R}(y_{\text{pred}})$ and we do not assume access to this at any point in training. 

\subsection{Group Relative Policy Optimization (GRPO)}
To optimize the policy, we adopt GRPO~\citep{shao2024deepseekmath}, a reinforcement learning algorithm that emphasizes relative rather than absolute performance. Instead of directly maximizing the reward of the current policy, GRPO evaluates the policy in relation to a group of baseline policies. This comparison helps improve learning stability, especially in settings with noisy or unsupervised reward signals.

Let \( \pi \) denote the current policy, and let \( \{\pi_1, \pi_2, \dots, \pi_K\} \) be a fixed or evolving set of reference policies. The GRPO objective is defined as:
\[
\mathcal{L}(\pi) = \mathbb{E}_{y_{\text{pred}} \sim \pi(x)} \left[ \mathcal{R}(y_{\text{pred}}) \right] - \frac{1}{K} \sum_{i=1}^{K} \mathbb{E}_{y_{\text{pred}} \sim \pi_i(x)} \left[ \mathcal{R}(y_{\text{pred}}) \right]
\]
The first term represents the expected reward under the current policy \( \pi \), while the second term computes the average reward across the reference group. The learning signal is thus the improvement in reward relative to these baselines. For more details, we refer the reader to~\citet{shao2024deepseekmath}.

\subsection{Entropy Reward}
For a given prompt \( x \), the model generates a response 
\( y_{\text{pred}} = y_{\text{pred},1}, \cdots, y_{\text{pred},T} = \pi(x) \), 
where \( T \) is the number of tokens in the response. At each token \( t \in \{1, \dots, T\} \), 
the model outputs a probability distribution \( p_t \) over the vocabulary \( \mathcal{V} \), i.e., 
\( p_t(v) = P(y_t = v \mid x, y_{<t}) \). The entropy of this distribution measures the model's uncertainty in predicting the next token and is given by:
\[
H(p_t) = -\sum_{v \in \mathcal{V}} p_t(v) \log p_t(v)
\]

To compute the total entropy of the response, we average the entropies across all tokens. The total entropy \( H(\pi(x)) \) provides a measure of the overall uncertainty in the model's response. 
Higher entropy indicates greater uncertainty or more diverse token predictions, while lower entropy 
suggests more confident and peaked distributions at each token. We use the negative entropy of the predicted token distribution as a reward signal: 
\[
\mathcal{R}(y_{\text{pred}}) = -H(\pi(x)) = \frac{1}{T}\sum_{t=1}^T \sum_{v \in \mathcal{V}} p_t(v) \log p_t(v)
\]
This reward encourages the model to produce more confident and peaked distributions over the vocabulary, effectively promoting lower uncertainty in its predictions. Within the RL framework, the learning objective becomes maximizing the expected reward over the data distribution:
\[
\max_\pi \mathbb{E}_{x \sim \mathcal{D}} \left[ \mathbb{E}_{y_{\text{pred}} \sim \pi(x)} \left[ \mathcal{R}(y_{\text{pred}}) \right] \right]
\]
By optimizing this objective, the model learns to generate responses with lower entropy without relying on external supervision or labeled target responses.

\begin{figure}
    \centering
    \includegraphics[width=0.98\textwidth]{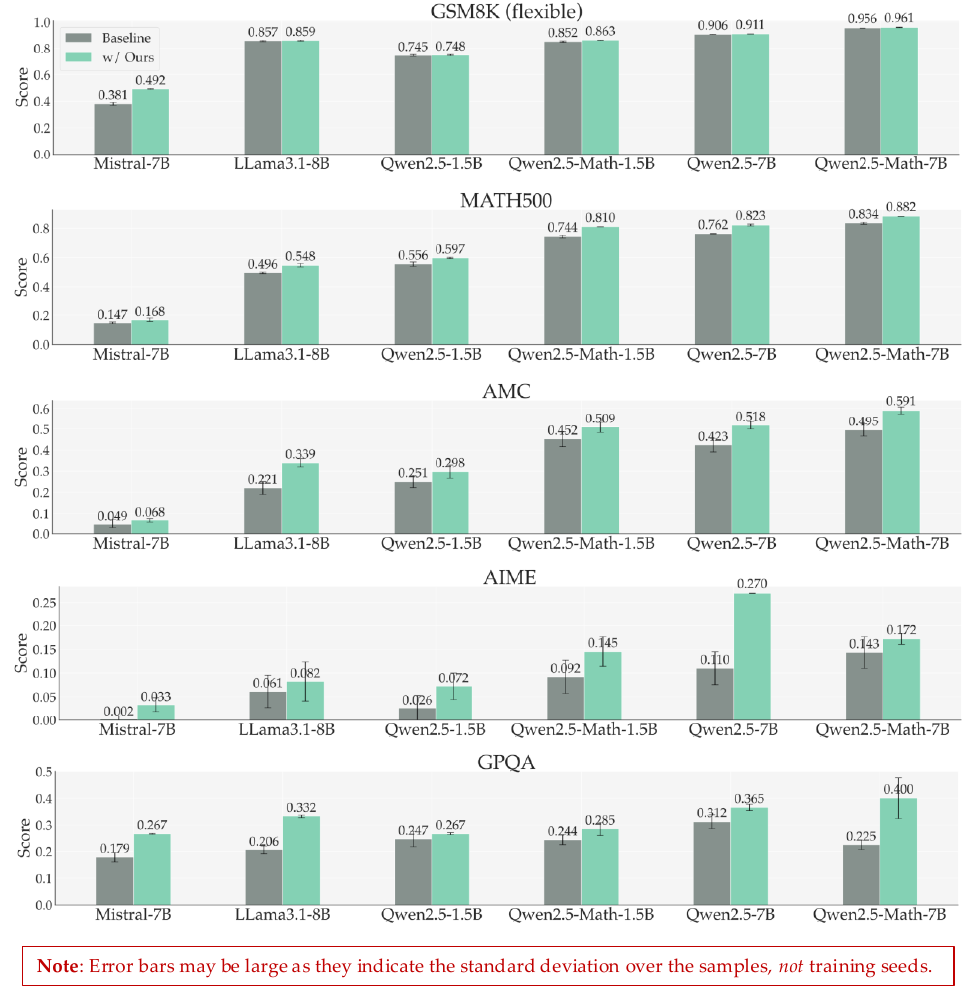}
    
    
    
    
    \caption{Performance on GSM8K, MATH500, AMC, AIME, and GPQA. The standard deviations reported are over 5, 5, 32, 64, and 10 samples, respectively. Across benchmarks and models, we find that entropy minimization alone is an effective reward for improving the reasoning ability of language models. All models are Instruct models; the "Instruct" is omitted for brevity.}\label{fig:main_fig}
\end{figure}

\section{Experiments}
\subsection{Experimental Setup}
\textbf{Benchmarks.} We train a model with reinforcement learning on each dataset independently. We conduct our experiments on the following commonly-used benchmarks for evaluating the reasoning capabilities of large language models: 
\begin{itemize}
    \item GSM8K~\citep{cobbe2021training}: GSM8K contains 8792 grade-school math word problems. The train set contains roughly 7473 problems and the test set contains roughly 1319 problems. 
    \item MATH500~\citep{hendrycks2021measuring,lightman2023let}: MATH~\citep{hendrycks2021measuring} is a dataset containing competition math problems spanning seven categories. It contains 12500 problems, of which 7500 are used for training and 5000 are used for testing. MATH500~\citep{lightman2023let} is a subset of the MATH test set created by OpenAI by sampling uniformly at random from the test set. 
    \item AMC~\citep{li2024numinamath}: The American Mathematics Competitions (AMCs) are competitions given to high school students. The specific dataset we use is comprised of 83 problems from the 2022 and 2023 AMC12 exams, which are given to 12th grade students. Although the original problems are in multiple-choice format, the dataset presents modified versions of the problem which expect an integer solution.
    \item AIME24~\citep{li2024numinamath}: The American Invitational Mathematics Examination (AIME) is a prestigious high school mathematics competition. It consists of 15 questions meant to be completed in 3 hours and is given to top-scoring students on the AMC exam. Each year, there are two versions of the exam which consist of distinct questions. We train on the 30 problems from both versions of the 2024 exam. 
    \item GPQA~\citep{rein2024gpqa}: GPQA is a dataset of 448 multiple-choice problems in biology, physics, and chemistry at the PhD level. They are intended to be "Google-proof" in the sense that they require advanced reasoning skills.
\end{itemize}
Since we are interested in test-time adaptation, and we do not assume access to the ground-truth answer, we use the same dataset for both training and evaluation. Additionally, some of the benchmarks do not have standardized train sets. The exception is GSM8K, where we use the standard train and test sets; this shows that generalization does occur and RENT is not merely overfitting to the test set.

\begin{figure}
    \centering
    \includegraphics[width=0.98\textwidth]{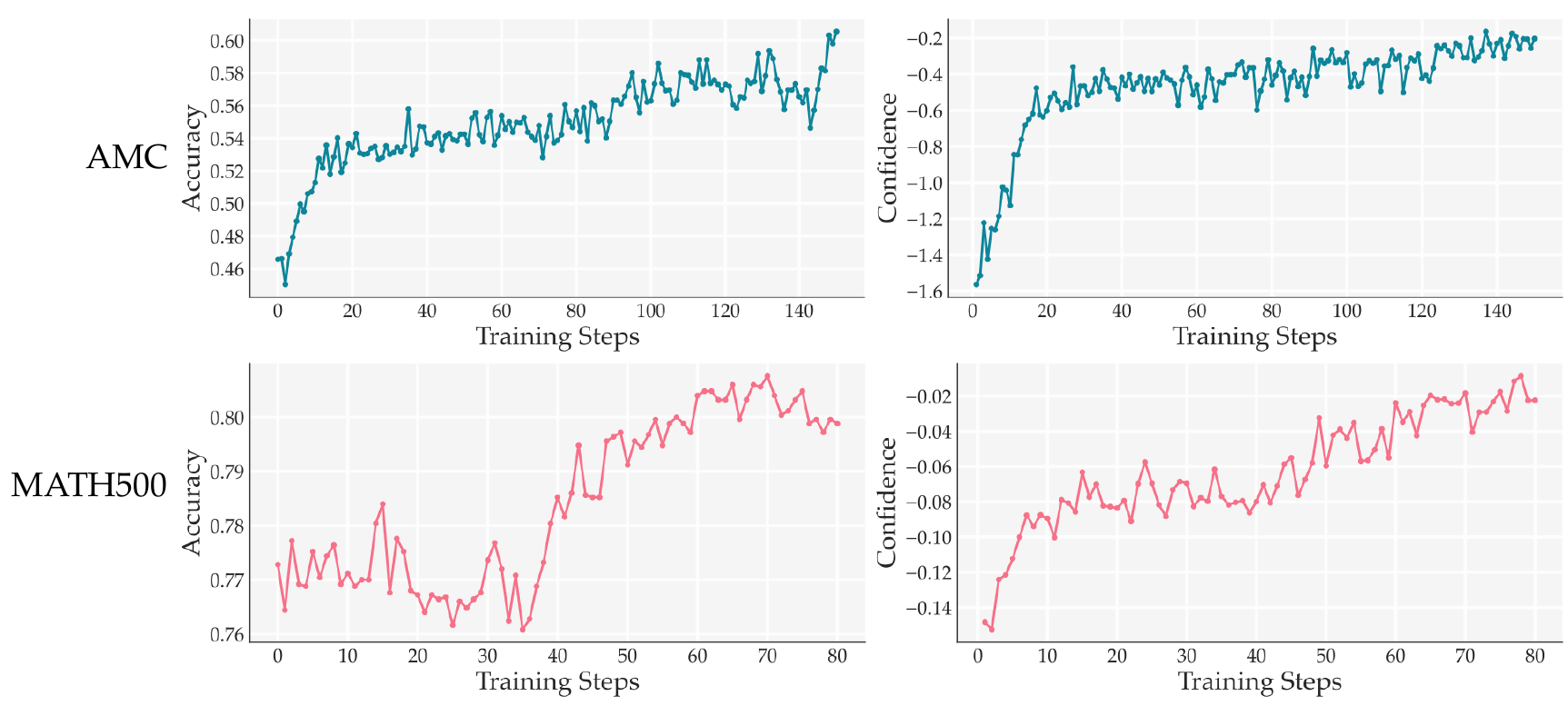}
    \caption{Accuracy and confidence over the course of training. The trends indicate that accuracy and confidence are indeed highly correlated and therefore it is natural to use confidence as a reward.}
    \label{accuracy_reward_graphs}
\end{figure}

\textbf{Models.} To showcase the generality of our method, we conduct experiments on a wide range of models from different model families and of varying parameter counts. We test on Mistral-7B-Instruct-v0.3, Llama3.1-8B-Instruct, Qwen2.5-1.5B-Instruct, Qwen2.5-Math-1.5B-Instruct, Qwen2.5-7B-Instruct, and Qwen2.5-Math-7B-Instruct.

\textbf{Implementation details.} For the RL optimization we use GRPO~\citep{shao2024deepseekmath} with a learning rate of $1 \times 10^{-6}$ and the Adam optimizer. The batch sizes and sampling hyperparameters may vary among models and datasets. We provide a full list of hyperparameters in the Appendix. 

\subsection{Main Results}
Figure~\ref{fig:main_fig} shows the performance of models before and after entropy minimization on GSM8K, MATH500, AMC, AIME24, and GPQA. We report standard deviations reported are over 5, 5, 32, 64, and 10 samples, respectively. Note that all models are Instruct models (e.g., Qwen2.5-1.5B refers to Qwen2.5-1.5B-Instruct). Across model families, model sizes, and benchmarks, entropy minimization allows large language models to improve their reasoning skills, without any external supervision. On the Math models such as Qwen2.5-Math-1.5B and Qwen2.5-Math-7B, the base model often struggles at following instructions and therefore the initial score is zero or near zero, and therefore the boost from entropy minimization is quite large. On models that are already proficient at instruction following, we can still see strong performance improvements from entropy minimization. Given the potential pitfall of overconfidence in language models, we performed extensive experimentation to ensure empirically that entropy minimization is a robust and generalizable reward function across datasets and models.

\subsection{Is It Just Formatting?}
It is a well-known issue with reasoning benchmarks that language models can lose points simply because they do not know how to put their answers in the right format. For example, MATH500 expects final answers to be placed in "boxed". A nontrivial amount of engineering effort has gone into both designing prompts that encourage correct formatting and implementing parsers that effectively extract answers from language model responses, in attempts to mitigate this issue. Therefore, one might wonder if, instead of learning to perform complex reasoning, RENT merely encourages the model to put its answers in the right format. Table \ref{format_table} shows that this is not the case. Models trained with the RENT reward outperform only using a format reward, which simply assigns a binary reward based on whether the correct format is followed in the response. In some cases, the performance of our method is similar to (or even slightly worse than) just using format reward, but of course it is expected that unsupervised RL methods might not always lead to significant improvements. For example, if the benchmark is extremely easy and the model only needs to learn the right format to achieve near-perfect scores, RENT would not outperform format reward. Or, if the benchmark is so hard that it is beyond the model's capabilities altogether, neither method would perform well. However, across datasets and model sizes, we find a consistent improvement over using the format reward and this assures us that the model is actually learning to think through difficult problems and improve its ability to reason. 

\begin{figure}
    \centering
    \includegraphics[width=0.98\textwidth]{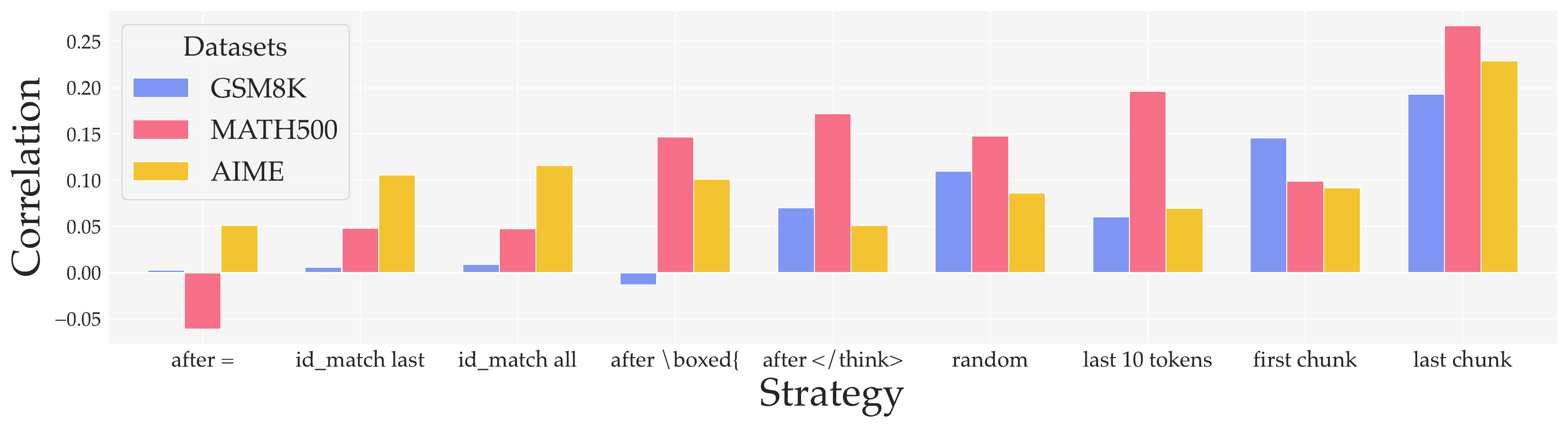}
    \caption{Evaluation (by computing correlation between accuracy and confidence) of various strategies for selecting which tokens to minimize the entropy over. We find the highest correlation between accuracy and confidence in the last few tokens of the response.}
    \label{strategy_bar}
\end{figure}

\begin{table}
\caption{Comparison to RL with a format reward. The best result on each benchmark is indicated in bold. RENT generally outperforms only using a format reward.}
\centering
\begin{tabular}{l@{\hskip 10pt}ccccc}
\toprule
 & GSM8K & MATH500 & AMC & AIME & GPQA \\
\midrule
\multicolumn{6}{l}{\textit{Mistral-7B-Instruct-v0.3}} \\
\quad Baseline & $0.381$ & $0.147$ & $0.049$ & $0.002$ & $0.179$ \\
\quad w/ Format reward only & $0.393$ & $0.150$ & $0.051$ & $0.015$ & $0.240$ \\ \quad w/ RENT (Ours) & $\mathbf{0.492}$ & $\mathbf{0.168}$ & $\mathbf{0.068}$ & $\mathbf{0.033}$ & $\mathbf{0.267}$ \\
\midrule
\multicolumn{6}{l}{\textit{LLama3.1-8B-Instruct}} \\
\quad Baseline & $0.857$ & $0.496$ & $0.221$ & $0.061$ & $0.206$ \\
\quad w/ Format reward only & $\mathbf{0.866}$ & $0.533$ & $0.265$ & $\mathbf{0.086}$ & $0.282$ \\ \quad w/ RENT (Ours) & $0.859$ & $\mathbf{0.548}$ & $\mathbf{0.339}$ & $0.082$ & $\mathbf{0.332}$ \\
\midrule
\multicolumn{6}{l}{\textit{Qwen2.5-1.5B-Instruct}} \\
\quad Baseline & $0.745$ & $0.548$ & $0.251$ & $0.026$ & $0.247$ \\
\quad w/ Format reward only & $\mathbf{0.754}$ & $0.558$ & $0.259$ & $0.054$ & $\mathbf{0.271}$ \\
 \quad w/ RENT (Ours) & $0.748$ & $\mathbf{0.597}$ & $\mathbf{0.298}$ & $\mathbf{0.072}$ & $0.267$ \\
\midrule
\multicolumn{6}{l}{\textit{Qwen2.5-Math-1.5B-Instruct}} \\
\quad Baseline & $0.852$ & $0.744$ & $0.452$ & $0.092$ & $0.244$ \\
\quad w/ Format reward only & $0.857$ & $0.756$ & $0.490$ & $0.117$ & $0.276$ \\
\quad w/ RENT (Ours) & $\mathbf{0.863}$ & $\mathbf{0.810}$ & $\mathbf{0.504}$ & $\mathbf{0.145}$ & $\mathbf{0.285}$ \\
\midrule
\multicolumn{6}{l}{\textit{Qwen2.5-7B-Instruct}} \\
\quad Baseline & $0.906$ & $0.762$ & $0.423$ & $0.110$ & $0.311$ \\
\quad w/ Format reward only & $\mathbf{0.913}$ & $0.774$ & $0.458$ & $0.156$ & $0.338$ \\
\quad w/ RENT (Ours) & $0.911$ & $\mathbf{0.823}$ & $\mathbf{0.518}$ & $\mathbf{0.270}$ & $\mathbf{0.365}$ \\
\midrule
\multicolumn{6}{l}{\textit{Qwen2.5-Math-7B-Instruct}} \\
\quad Baseline & $0.956$ & $0.834$ & $0.495$ & $0.143$ & $0.225$ \\
\quad w/ Format reward only & $0.957$ & $0.873$ & $0.560$ & $0.154$ & $0.340$ \\
\quad w/ RENT (Ours) & $\mathbf{0.967}$ & $\mathbf{0.882}$ & $\mathbf{0.591}$ & $\mathbf{0.167}$ & $\mathbf{0.400}$ \\
\bottomrule
\end{tabular}
\label{format_table}
\end{table}

\subsection{Correlation Between Entropy and Accuracy}
Figure \ref{accuracy_reward_graphs} shows the accuracy and confidence throughout training Qwen2.5-Math-7B and Qwen2.5-7B-Instruct on the AMC and MATH500 datasets respectively. Critically, as the model improves its confidence via RENT, the accuracy of the model improves as well. This demonstrates the significant correlation between answer confidence and answer accuracy, supporting our initial hypothesis.

\subsection{Comparison to Concurrent Work}
In this section, we compare RENT to concurrent papers which use intrinsic rewards. We evaluate on GSM8K, MATH500, AMC, AIME, and GPQA and run all experiments with Qwen2.5-7B-Instruct as the baseline model. Table~\ref{ttrl_comp} shows comparisons to the following methods:
\begin{itemize}
    \item Test-Time Reinforcement Learning (TTRL)~\citep{zuo2025ttrl} assigns a reward of 1 to the majority answer and 0 to all other answers. In our experiments, we reimplemented this majority voting reward in our codebase. 
    \item Intuitor~\citep{zhao2025learning} uses the forward KL divergence between a uniform distribution and the model's distribution as the reward. In contrast, we use entropy, which is the reverse KL divergence from the uniform distribution. Intuitor is mode-seeking while RENT is mode-covering. We ran the publicly available Intuitor code (which is also implemented on top of verl framework \cite{sheng2024hybridflow}) with the same batch size, epochs and evaluation strategy as RENT for fair comparison. 
    \item \citet{shao2025spurious} suggested that even random or "spurious" rewards could be used to improve reasoning. To compare against spurious rewards, we modify our code to set the reward for every generation randomly to 0 or 1 with equal probability. Intuitively, we believe spurious rewards might work because gradients from correct examples contribute to learning, while gradients from incorrect examples might cancel each other out. Our hypothesis is supported by work such as ~\citet{rolnick2017deep}, which shows that learning can still happen even when diluting datasets with incorrect labels.
\end{itemize}
Empirically, we find that RENT is the best of the four methods on average. Compared to TTRL and Intuitor, performance is similar on most benchmarks except AIME, where RENT outperforms both by a large margin. This is especially interesting since AIME is the hardest benchmark in our evaluations (i.e., the initial accuracy of the model is the lowest). Spurious rewards are not competitive with the other three methods; we conclude that random reward values are not enough and it is indeed beneficial to use meaningful unsupervised rewards that explicitly encourage some measure of confidence.

\begin{table}[h!]
\caption{Comparison of RENT with three concurrent papers: TTRL~\citep{zuo2025ttrl}, Intuitor~\citep{zhao2025learning}, and Spurious Rewards~\citep{shao2025spurious}. The best result on each benchmark is indicated in bold. RENT is the best-performing method on MATH500, AIME, GPQA, and is the best on average.}
\centering
\begin{tabular}{lccccc|c}
\toprule
 & GSM8K & MATH500 & AMC & AIME & GPQA & Average \\
\midrule
\textit{Qwen2.5-7B-Instruct} & & & & & \\
\quad w/ TTRL & \textbf{0.933} & 0.822 & 0.521 & 0.172 & 0.346 & 0.559 \\
\quad w/ Intuitor (forward KL) & 0.929 & 0.783 & \textbf{0.525} & 0.200 & 0.337 & 0.555 \\
\quad w/ Spurious Rewards & 0.910 & 0.774 & 0.459 & 0.156 & 0.342 & 0.528 \\
\quad w/ RENT (Ours) & 0.911 & \textbf{0.823} & 0.518 & \textbf{0.270} & \textbf{0.365} & \textbf{0.577} \\
\bottomrule
\end{tabular}
\label{ttrl_comp}
\end{table}

\subsection{Qualitative Samples}
Table \ref{tab:gsm_qual} shows a qualitative sample from GSM8K and Table \ref{tab:aime_qual} shows a qualitative sample from AIME. The qualitative samples verify that the model indeed learns meaningful reasoning skills via entropy minimization. It is not merely learning to format its answer correctly, or otherwise collapsing to some other reward-hacking behavior. 

\begin{table}[h!] 
\caption{The model splits the problem into its key steps and correctly calculates the target age}
\setlength{\tabcolsep}{10pt} 
\renewcommand{\arraystretch}{1.5} 
\centering
\begin{tabular}{|l|p{11cm}|}
\hline
\textbf{Prompt} & 
system

Please reason step by step, and put your final answer within \textbackslash boxed\{\}.
user

Jaymee is 2 years older than twice the age of Shara. If Shara is 10 years old, how old is Jaymee? Let's think step by step and output the final answer within "\textbackslash boxed\{\}".

assistant\\
\hline
\textbf{Response} &
To determine Jaymee's age, we need to follow these steps:

1. Identify Shara's age.

2. Calculate twice Shara's age.

3. Add 2 years to twice Shara's age.

Step 1: Shara's age is given as 10 years old.

Step 2: Twice Shara's age is calculated as:
\[ 2 \times 10 = 20 \]

Step 3: Jaymee is 2 years older than twice Shara's age, so we add 2 years to 20:
\[ 20 + 2 = 22 \]

Therefore, Jaymee is \boxed{22} years old.\\
\hline
\end{tabular}
\label{tab:gsm_qual}
\end{table}

\begin{table}[h!]
\caption{The model identifies the geometric series and succinctly soles the problem afterwards}
\setlength{\tabcolsep}{10pt} 
\renewcommand{\arraystretch}{1.5} 
\centering
\begin{tabular}{|l|p{11cm}|}
\hline
\textbf{Prompt} & 
system

Please reason step by step, and put your final answer within \textbackslash boxed\{\}.

user

The infinite product
\[\sqrt[3]{10} \cdot \sqrt[3]{\sqrt[3]{10}} \cdot \sqrt[3]{\sqrt[3]{\sqrt[3]{10}}} \cdots\]
evaluates to a real number $\sqrt{m}$. What is the value of m? Let's think step by step and output the final answer within \textbackslash boxed\{\}.

assistant\\
\hline
\textbf{Response} &
To find the value of the infinite product \(\sqrt[3]{10} \cdot \sqrt[3]{\sqrt[3]{10}} \cdot \sqrt[3]{\sqrt[3]{\sqrt[3]{10}}} \cdots\), we can express each term in the product as a power of 10.

The first term is \(\sqrt[3]{10} = 10^{1/3}\).
The second term is \(\sqrt[3]{\sqrt[3]{10}} = \left(10^{1/3}\right)^{1/3} = 10^{1/9}\).
The third term is \(\sqrt[3]{\sqrt[3]{\sqrt[3]{10}}} = \left(10^{1/9}\right)^{1/3} = 10^{1/27}\).
And so on.

The infinite product can be written as:
\[10^{1/3} \cdot 10^{1/9} \cdot 10^{1/27} \cdots = 10^{1/3 + 1/9 + 1/27 + \cdots}.\]

The exponent is an infinite geometric series with the first term \(a = \frac{1}{3}\) and common ratio \(r = \frac{1}{3}\). The sum \(S\) of an infinite geometric series is given by:
\[S = \frac{a}{1 - r} = \frac{\frac{1}{3}}{1 - \frac{1}{3}} = \frac{\frac{1}{3}}{\frac{2}{3}} = \frac{1}{2}.\]

So the infinite product evaluates to:
\[10^{1/2} = \sqrt{10}.\]

Thus, the value of \(m\) is 10. The final answer is:
\[
\boxed{10}
\]\\
\hline
\end{tabular}
\label{tab:aime_qual}
\end{table}

\subsection{Which Tokens to Minimize Entropy Over}
We investigated which response tokens are the best to minimize entropy over. To do this, we evaluated the following response token selection strategies:
\begin{itemize}
    \item \textbf{last chunk} - split the response into $k$ chunks, and select all tokens in the last chunk
    \item \textbf{first chunk} - split the response into $k$ chunks, and select all tokens in the first chunk
    \item \textbf{last 10 tokens} - select the last $k$ response tokens
    \item \textbf{random} - select $k$ random tokens from the response
    \item \textbf{after </think>} - select all tokens after the thought process (after the "</think>" tag)
    \item \textbf{after \textbackslash boxed\{} - select all tokens after the boxed answer
    \item \textbf{id\_match\_last} - select the tokens in the model's final answer
    \item \textbf{id\_match\_all} - select all tokens tokens in the model's final answer and any previous occurrences
    \item \textbf{after =} - minimize all tokens after the equality symbol ("=")
\end{itemize}

Figure \ref{strategy_bar} shows the initial correlation between negative entropy and accuracy for each of these strategies on three datasets. While most token selection strategies do result in a positive correlation between confidence and accuracy, we note that the "last chunk" strategy has a significantly higher correlation compared to the "first chunk" strategy. This suggests that the most important tokens to minimize entropy over are tokens that occur later in the response. Furthermore, based on the low correlation results from the "id\_match\_last" and "id\_match\_all" strategies, we find that it is not sufficient to simply minimize the entropy of the final answer tokens; this suggests that, counterintuitively, the token-level confidence of the final answer tokens is not well-calibrated to its true response confidence/accuracy.

\section{Limitations}
Fundamentally, unsupervised learning alone is relatively limited compared to methods which are able to use external supervision for learning. Therefore, it is not surprising that our method is not able to match the performance of methods that have access to the ground-truth answers. It is, of course, a possibility for the model to be confidently wrong. Overconfidence is a well-known issue with language models and these calibration errors can cause RENT to fail catastrophically. It could be dangerous to deploy such an unsupervised learning method in the real world without any safeguards. However, we generally find empirically that confidence does correlate with accuracy and the performance does improve by using confidence alone. This indicates that even if the model is overconfident on some answers, it is well-calibrated overall.

\section{Conclusion}
We presented RENT, an unsupervised reinforcement learning method which uses entropy as a reward. In our experiments, we showed that by simply minimizing entropy, we can improve the reasoning performance of language models on GSM8K, MATH500, AMC, AIME, and GPQA. Our reward function is general and can be applied on a wide range of domains. We are excited about the possibility of using entropy minimization and, more broadly, unsupervised reinforcement learning to improve the capabilities of machine learning models in regimes where external supervision is unavailable. 

\section{Acknowledgements}
We would like to acknowledge Nikhil Chandak, Shashwat Goel, and Ameya Prabhu for insights on evaluation that helped us to improve our paper.

\bibliographystyle{plainnat}
\bibliography{main}

\newpage
\appendix
\section{Hyperparameters}
A full list of hyperparameters can be found in Table~\ref{tbl:hyperparameters}.
\begin{table}[ht]
\caption{Hyperparameters.}\label{tbl:hyperparameters}
\vskip 0.15in
\begin{center}
\begin{small}
\begin{tabular}{ll}
\toprule
\textbf{Hyperparameter} & \textbf{Value}  \\
\midrule
Max prompt length & 1024 \\
Max response length & 3072 \\
Batch size & 64 GSM8K
        \\ & 500 MATH500
        \\ & 80 AMC
        \\ & 30 AIME
        \\ & 64 Countdown
        \\ & 196 GPQA \\
Policy mini batch size & 32 GSM8K
                     \\ & 32 MATH500
                     \\ & 80 AMC
                     \\ & 30 AIME
                     \\ & 32 Countdown
                     \\ & 32 GPQA \\
Policy micro batch size per GPU & 8 \\
Learning rate & $1 \times 10^{-6}$ \\
Weight decay & 0.01 \\
Learning rate warmup & Constant \\
Optimizer & Adam \\
Temperature & 1.0 for train \\ & 0.8 for validation \\
Top $k$ & -1 \\
Top $p$ & 1 \\
Number of samples per example $n$ & 5 \\
Remove padding & True \\
Use KL loss & True \\
KL loss coefficient & 0.001 \\
Clip ratio & 0.2 \\
Grad clip & 1.0 \\
\bottomrule
\end{tabular}
\end{small}
\end{center}
\end{table}

\end{document}